\documentclass[sigconf,natbib=true,anonymous=False]{acmart}

\AtBeginDocument{%
  \providecommand\BibTeX{{%
    Bib\TeX}}}
\copyrightyear{2023}
\acmYear{2023}
\setcopyright{acmlicensed}\acmConference[CIKM '23]{Proceedings of the 32nd ACM International Conference on Information and Knowledge Management}{October 21--25, 2023}{Birmingham, United Kingdom}
\acmBooktitle{Proceedings of the 32nd ACM International Conference on Information and Knowledge Management (CIKM '23), October 21--25, 2023, Birmingham, United Kingdom}
\acmPrice{15.00}
\acmDOI{10.1145/3583780.3614766}
\acmISBN{979-8-4007-0124-5/23/10}

\usepackage{amsmath,multirow,booktabs,tabularx}
\usepackage{array} 
\usepackage{longtable}
\usepackage{booktabs}
\usepackage{float}
\usepackage{anyfontsize}
\usepackage{color} 
\usepackage{algorithm}  
\usepackage{algorithmicx}  
\usepackage{algpseudocode}  
\usepackage{amsmath} 
\usepackage{amsfonts}
\usepackage{graphicx}
\usepackage{bm}
\usepackage{subfigure}
\usepackage{parskip}
\usepackage{multirow}
\usepackage{bigstrut}
\usepackage{balance} 
\def\mc{\mathcal}

\def\bs{\boldsymbol}
\usepackage{balance} 

\begin{document}

\title{A Multi-Task Semantic Decomposition Framework with Task-specific Pre-training for Few-Shot NER}
\author{Guanting Dong}
\authornote{The first three authors contribute equally.}
\email{dongguanting@bupt.edu.cn}

\affiliation{%
  \institution{Beijing University of Posts and Telecommunication}
  \city{Beijing}
  \country{China}
}

\author{Zechen Wang}
\email{shenshui@bupt.edu.cn}
\authornotemark[1]
\affiliation{%
  \institution{Beijing University of Posts and Telecommunication}
  \city{Beijing}
  \country{China}
}

\author{Jinxu Zhao}
\email{zhaojinxu@bupt.edu.cn}
\authornotemark[1]
\affiliation{%
  \institution{Beijing University of Posts and Telecommunication}
  \city{Beijing}
  \country{China}
}

\author{Gang Zhao}
\email{zhaogang@bupt.edu.cn}
\affiliation{%
  \institution{Beijing University of Posts and Telecommunication}
  \city{Beijing}
  \country{China}
}

\author{Daichi Guo}
\email{guodaichi@bupt.edu.cn}
\affiliation{%
  \institution{Beijing University of Posts and Telecommunication}
  \city{Beijing}
  \country{China}
}

\author{Dayuan Fu}
\email{fdy@bupt.edu.cn}
\affiliation{%
  \institution{Beijing University of Posts and Telecommunication}
  \city{Beijing}
  \country{China}
}

\author{Tingfeng Hui}
\email{huitingfeng@bupt.edu.cn}
\affiliation{%
  \institution{Beijing University of Posts and Telecommunication}
  \city{Beijing}
  \country{China}
}

\author{Chen Zeng}
\email{chenzeng@bupt.edu.cn}
\affiliation{%
  \institution{Beijing University of Posts and Telecommunication}
  \city{Beijing}
  \country{China}
}

\author{Keqing He}
\email{hekeqing@meituan.com}
\affiliation{%
  \institution{Meituan Group, Beijing}
  \city{Beijing}
  \country{China}
}

\author{Xuefeng Li}
\email{lixuefeng@bupt.edu.cn}
\affiliation{%
  \institution{Beijing University of Posts and Telecommunication}
  \city{Beijing}
  \country{China}
}

\author{Liwen Wang}
\email{w_liwen@bupt.edu.cn}
\affiliation{%
  \institution{Beijing University of Posts and Telecommunication}
  \city{Beijing}
  \country{China}
}

\author{Xinyue Cui}
\email{tracycui@bupt.edu.cn}
\affiliation{%
  \institution{Beijing University of Posts and Telecommunication}
  \city{Beijing}
  \country{China}
}

\author{Weiran Xu}
\email{xuweiran@bupt.edu.cn}
\authornote{Weiran Xu is the corresponding author}
\affiliation{%
  \institution{Beijing University of Posts and Telecommunication}
  \city{Beijing}
  \country{China}
}

\renewcommand{\shortauthors}{Guanting Dong et al.}

\begin{abstract}
 The objective of few-shot named entity recognition is to identify named entities with limited labeled instances. Previous works have primarily focused on optimizing the traditional token-wise classification framework, while neglecting the exploration of information based on NER data characteristics. To address this issue, we propose a \textbf{M}ulti-Task \textbf{S}emantic \textbf{D}ecomposition Framework via Joint Task-specific \textbf{P}re-training (\textbf{MSDP}) for few-shot NER. Drawing inspiration from demonstration-based and contrastive learning, we introduce two novel pre-training tasks: Demonstration-based Masked Language Modeling (MLM) and Class Contrastive Discrimination. These tasks effectively incorporate entity boundary information and enhance entity representation in Pre-trained Language Models (PLMs). In the downstream main task, we introduce a multi-task joint optimization framework with the semantic decomposing method, which facilitates the model to integrate two different semantic information for entity classification. Experimental results of two few-shot NER benchmarks demonstrate that MSDP consistently outperforms strong baselines by a large margin. Extensive analyses validate the effectiveness and generalization of MSDP.
\end{abstract}

\ccsdesc[500]{Computing methodologies~Artificial intelligence}
\ccsdesc[300]{Natural language processing~Information extraction}

\keywords{Few-shot NER, Multi-Task, Semantic Decomposition, Pre-training}



\maketitle

\section{Introduction}
Named entity recognition (NER) plays a crucial role in Natural Language Understanding applications by identifying consecutive segments of text and assigning them to predefined categories \cite{molla2006named,nadeau2007survey,guo2009named}. Recent advancements in deep neural architectures have demonstrated exceptional performance in fully supervised NER tasks \cite{lample2016neural,chiu2016named,peters2017semi}. However, the collection of annotated data for practical applications incurs significant expenses and poses inflexibility challenges. As a result, the research community has increasingly focused on few-shot NER task, which seeks to identify entities with only a few labeled instances, attracting substantial interest in recent years.

Previous few-shot NER methods \cite{yang2020simple,hou2020few,ziyadi2020example,das2021container,fritzler2019few,ma2022label} generally formulate the task as a sequence labeling task based on prototypical networks \cite{snell2017prototypical}. These approaches employ prototypes to represent each class based on labeled instances and utilize the nearest neighbor method for NER. However, these models only capture the surface mapping between entity and class, making them vulnerable to disturbances caused by non-entity tokens (i.e. "O" class) \cite{wang2021enhanced,shen2021locate}. To alleviate this issue, a branch of two-stage methods \cite{ma2022decomposed,wang2021enhanced,shen2021locate,wu2022propose} arise to decouple NER into two separate processes, including span extraction and entity classification. Despite the above achievement
\begin{figure}[t]
\centering

\resizebox{.47\textwidth}{!}{\includegraphics{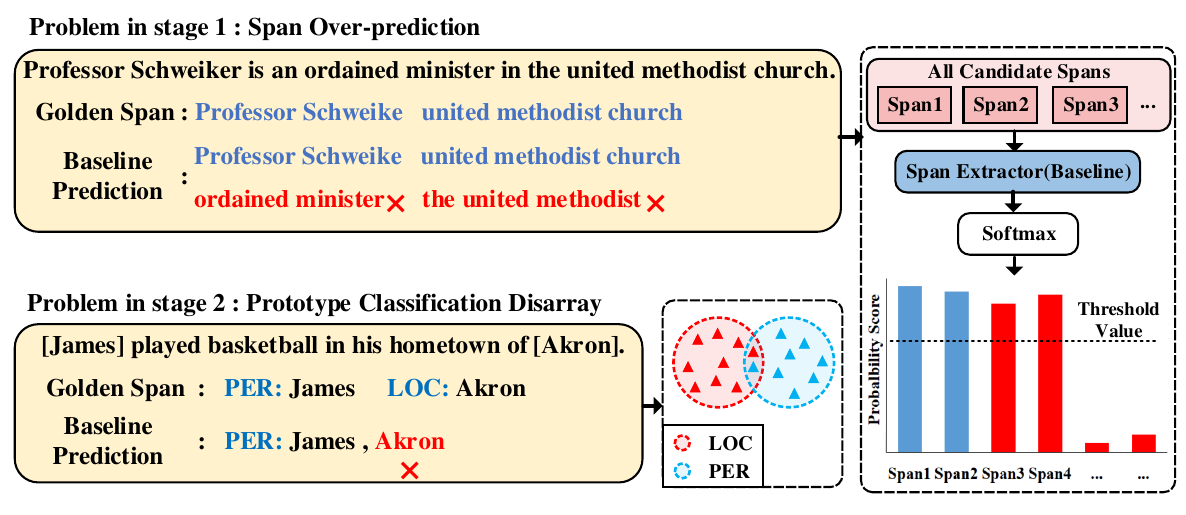}}

\vspace{-0.2cm} 
\caption{The illustration of the baseline model suffering from span over-prediction (upper) and Prototype Classification Disarray (down) problem in few-shot NER.}
\label{fig:intro}
\vspace{0.2cm} 
\end{figure}
, there are still two remaining problems. (1) \textbf{Span Over-prediction:} as shown in Figure \ref{fig:intro} , previous span-based works suffer from the span over-prediction problem \cite{zhu2022boundary,he2020contrastive}. Specifically, the model will extract redundant candidate spans in addition to predicting the correct spans. The reason for the above phenomenon is that it is difficult for PLMs to learn reliable boundary information between entities and non-entities due to insufficient data. As a result, PLMs tend to give similar candidate spans high probability scores or even be over-confident about their predictions \cite{guo2017calibration}.
(2) \textbf{Prototype Classification Disarray:} Previous prototype-based methods directly utilize the mean value of entity representations to compute prototype embedding, leading to the classification accuracy heavily relies on the quality of entity representations. Unfortunately, PLMs often face the issue of semantic space collapse, where different classes of entity representations are closely distributed, especially for entities within the same sentences. Figure \ref{fig:intro} illustrates that different classes of entities interfere with each other under the interaction of the self-attention mechanism, causing close or even overlapping prototypes distribution in the semantic space(e.g. "LOC" prototype overlapping with “PER” prototype). The model finally suffers from performance degradation due to class confusion. Therefore, we urgently need to design a method introducing different aspects of information to alleviate the above problems, which facilitates techniques of few-shot NER to be widely applied in realistic task-oriented dialogue scenarios.

To tackle these limitations,  we propose a \textbf{M}ulti-Task \textbf{S}emantic 
\textbf{D}ecomposition Framework via Joint Task-specific \textbf{P}re-training (\textbf{MSDP}), which guides PLMs to capture reliable entity boundary information and better entity representations of different classes. For the pre-training stage, inspired by demonstration-based learning \cite{gao2020making} and contrastive learning \cite{chen2020simple}, we introduce two novel task-specific pre-training tasks according to the data characteristics of NER (entity-label pairs):
\textbf{Demonstration-based MLM}, in which we design three kinds of demonstrations containing entity boundary information and entity label pair information. PLMs will implicitly learn the above information during predicting label words for [MASK];
\textbf{Class Contrastive Discrimination}, in which we use contrastive learning to better discriminate different classes of entity representations by constructing positive, negative, and hard negative samples.
Through the joint optimization of above fine-grained pre-training tasks, PLMs can effectively alleviate the two remaining problems.

For downstream few-shot NER, we follow the two-stage framework \cite{ma2022decomposed,wang2022spanproto} including span extraction and entity classification, and initialize them with the pre-trained parameters. Different from previous methods, we employ a multi-task joint optimization framework and utilize different masking strategies to decompose class-oriented prototypes and contextual fusion prototypes. The purpose of our design is to assist the model to integrate different semantic information for classification, which further alleviates the prototype classification disarray problem. We conduct extensive experiments over two widely-used benchmarks, including Few-NERD \cite{ding2021few} and CrossNER \cite{hou2020few}. Results show that our method consistently outperforms state-of-the-art baselines by a large margin. In addition, we introduce detailed experimental analyses to further verify the effectiveness of our method.
\begin{figure*}
    \centering
    \resizebox{\textwidth}{!}{
    \includegraphics{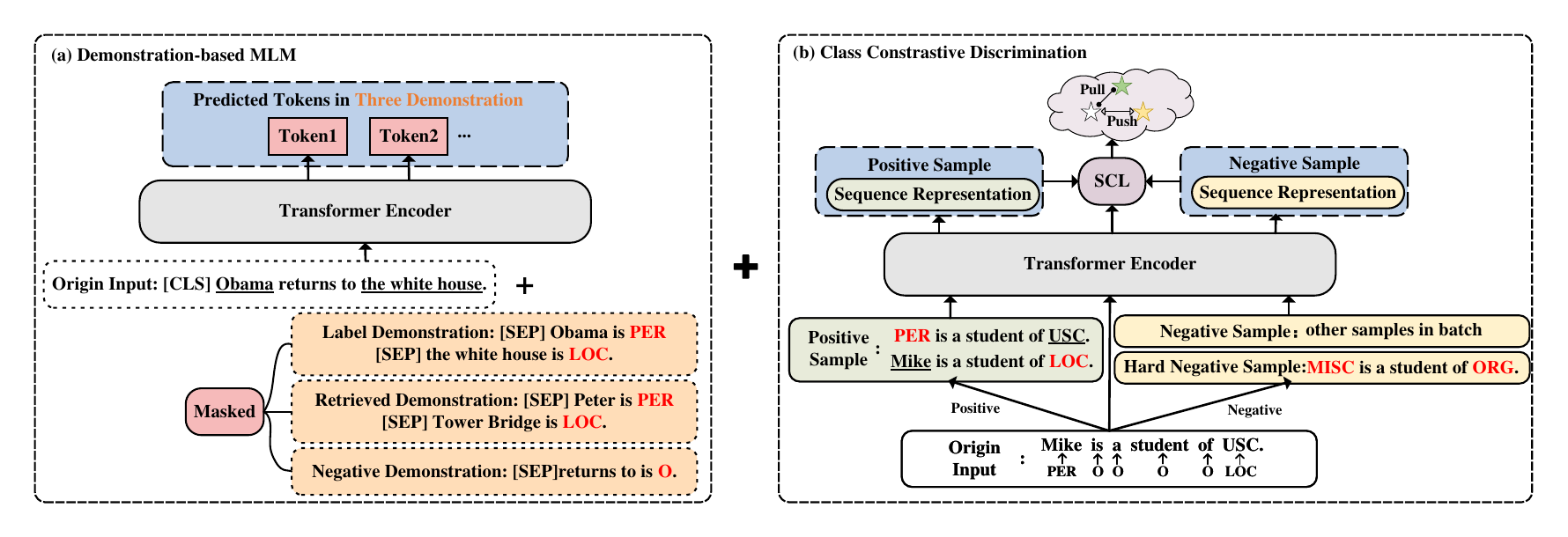}}
    \vspace{-0.8cm}
    \caption{The illustration of two task-specific pre-training tasks.}
    \label{fig:pretrain}
    \vspace{0.2cm}
\end{figure*}
Our contributions are three-fold: 

1)~To the best of our knowledge, we are the first to introduce a multi-task joint optimization framework with the semantic decomposing method into Few-Shot NER task.

2)~we futher propose two task-specific pre-training tasks via demonstration and contrastive learning, namely demonstration-based MLM and class contrastive discrimination, for effectively injecting entity boundary information and better entity representation into PLMs.


3)~Experiments on two widely-used few-shot NER benchmarks show that our framework achieves superior performance over previous state-of-the-art methods. Extensive analyses further validate the effectiveness and generalization of MSDP. Our source codes and datasets are available at Github\footnote{https://github.com/dongguanting/MSDP-Fewshot-NER} for further comparisons.

\section{Related work}

\subsection{Few-shot NER.}
Few-shot NER aims to enhance the performance of model identifying and classifying entities with only little annotated data \cite{ghaddar2018transforming,cao2019low,liu2020coach,jia2019cross,liu2020zero,liu2021importance,jia2020multi,zhao-etal-2022-entity,liu2021crossner,10193387,li-etal-2023-generative}. For few-shot NER, a series of approaches have been proposed to learn the representation of entities in the semantic space, i.e. prototypical learning \cite{snell2017prototypical}, margin-based learning \cite{levi2021rethinking} and contrastive learning \cite{gao2021simcse,10094766,9747192}. Existing approaches can be divided into two kinds, i.e., one-stage \cite{snell2017prototypical,hou2020few,das2021container,ziyadi2020example}  and two-stage \cite{ma2022decomposed,wu2022propose,shen2021locate,10095149}. Generally, the methods in the kind of one-stage typically categorize the entity type by token-level metric learning. In contrast, two-stage mainly focuses on two training stages consisting of entity span extraction and mention type classification.

\subsection{Task-specific pre-training Models.}
Pre-trained language models have been applied as an integral component in modern NLP systems for effectively improving downstream tasks \cite{peters1802deep,radford2019language,zhang-etal-2023-pay,devlin2018bert,yang2019xlnet,liu2019roberta,qixiang-etal-2022-exploiting,zeng2022semisupervised}. Due to the underlying discrepancies between the language modeling and downstream tasks, task-specific pre-training methods have been proposed to further boost the task performance, such as SciBERT \cite{beltagy2019scibert}, VideoBERT \cite{sun2019videobert}, DialoGPT \cite{zhang2019dialogpt}, PLATO \cite{bao2019plato}, Code-BERT \cite{feng2020codebert}, ToD-BERT \cite{wu2020tod} and VL-BERT \cite{su2019vl}. However, most studies in the field of few-shot NER use MLM and other approaches for Data Augmentation \cite{hou2018sequence,zhou2022melm,dong2022pssat}. Although DictBERT \cite{chen2022dictbert}, NER-BERT \cite{liu2021ner} and others have conducted pre-training, their methods are too generalized to adapt to the structured data features of NER or propose optimization for specific problems. Therefore, we designed demonstration-based learning pre-training and contrastive learning pre-training for NER tasks to improve the performance of the model.

\subsection{Demonstration-based learning}
\label{sec:demonstration}
Demonstrations are first introduced by the GPT series \cite{radford2019language,brown2020language}, where a few examples are sampled from training data and transformed with templates into appropriately-filled prompts. Based on the task reformulation and whether the parameters are updated, the existing demonstration-based learning research can be broadly divided into three categories: In-context Learning \cite{brown2020language,zhao2021calibrate,min2021noisy,wei2022chain}, Prompt-based Fine-tuning \cite{liang2022contrastive}, Classifier-based Fine-tuning \cite{lee2021good,yuan2023scaling}. However, these approaches mainly adopt demonstration-based learning in the fine-tuning that cannot make full use of the effect of demonstration-based learning. Different from them, we use demonstration-based learning in the pre-training stage that can better capture the entity boundary information to solve the multiple-span prediction problem.

\begin{figure*}
    \centering
    \resizebox{.95\textwidth}{!}{
    \includegraphics{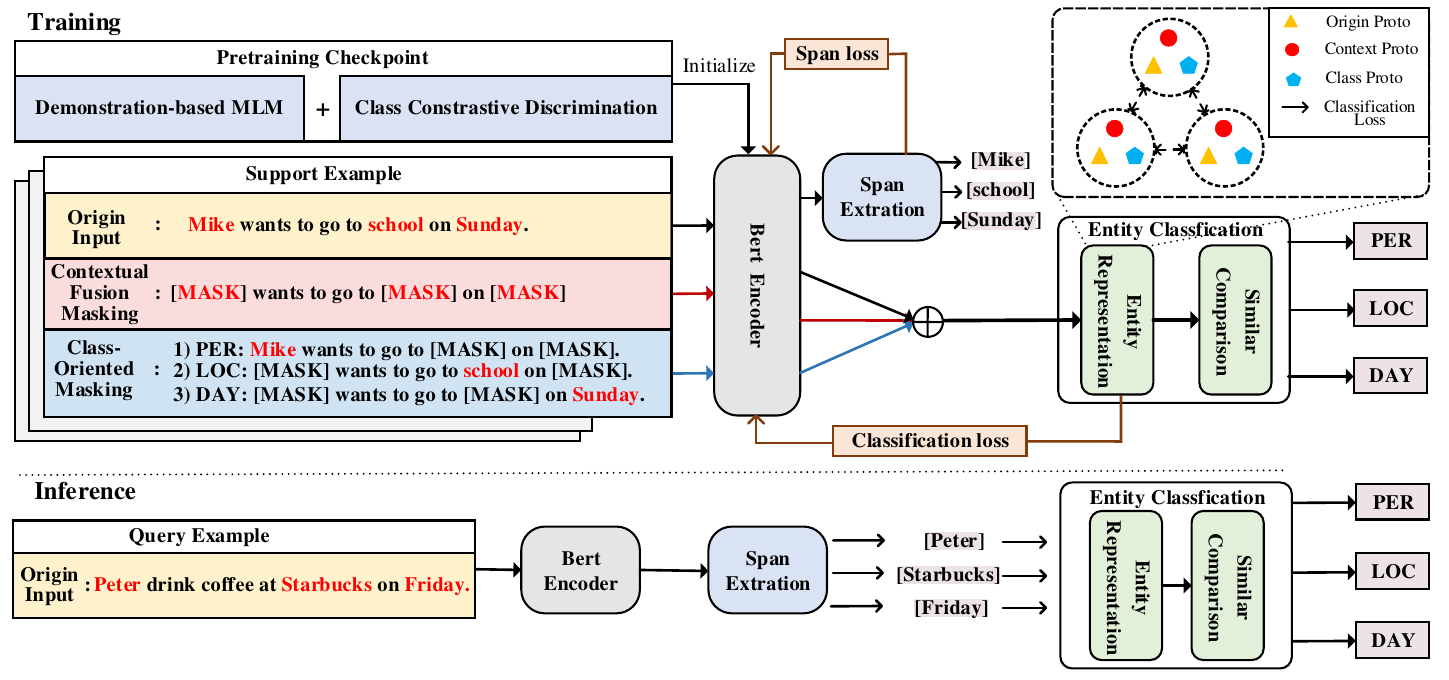}}
    \vspace{-0.2cm}
    \caption{The overall architecture of our proposed MSDP framework}
    \label{fig:main}
\end{figure*}

\section{Method}

\subsection{Task-Specific Pre-training}
The performance of few-shot NER depends heavily on the different aspects of information from entity label pairs. as shown in Figure \ref{fig:pretrain}, we introduce two novel pre-training tasks: 1) demonstration-based MLM and 2) contrastive entity discrimination, to learn the different aspects of knowledge.

\textbf{Demonstration-based MLM.} We follow the design of masked language modeling (MLM) in BERT \cite{devlin2018bert} and integrate the idea of demonstration-based learning on this basis. In order to prompt PLMs to figure out the boundary between the entity and none-entity, we propose three different demonstrations which are shown in Figure \ref{fig:pretrain}(a):

1)~Label demonstration (LD): We let $D_{train}$  denote the train dataset. For each input $x$ in $D_{train}$, we extract the entity label pair $(e, l)$ belonging to $x$ and then concatenate them behind input $x$ in form of the simple template $T= \{e \; is \; l\}$. Different demonstrations are separated by \emph{[SEP]}.

2)~Retrieved demonstration (RD): Given an entity type label set $L =\{l_{1}, . . . , l_{|L|}\}$, we first enumerate all the entities in $D_{train}$ and create a mapping $M=\{l_i : [e_1, . . . , e_n] \, \bm{|} \, l_i \in L\}$ between $l$ and the corresponding list of entities. Further, we randomly select $K$ entity label pairs $(e, l)$ from the mapping $M$ according to the label set $L_x$ appearing in input $x$, which aims at introducing rich entity label pair information to prompt the model. Furthermore, we concatenate them behind label demonstration with template $T= \{e \; is \; l\}$.

3) Negative demonstration (ND):  We randomly select $K$ none-\\entities that are easily confused by the model from input $x$ to construct negative sample pairs $(e_{none}, O)$, and then concatenate them behind retrieved demonstrations with template $T'= \{e_{none} \; is \; O\}$. Therefore, our training samples can be formulated as:
\begin{equation}
\begin{aligned}
\mathtt{[CLS]} \,  x \, \mathtt{[SEP]} \, LD \, \mathtt{[SEP]} \, RD \, \mathtt{[SEP]} \, ND
\end{aligned}
\end{equation}
After constructing the training set, we randomly randomly replace N entities or labels with mask symbols or labels in the demonstration with the special \emph{[MASK]} symbol\footnote{N is an empirical hyperparameter, which is set to 4.}, and then try to recover them. If entity $e$ consists of multiple tokens, all of the component tokens will be masked. Hence, the loss function of the MLM is:
\begin{equation}
\begin{aligned}
{L}_{mlm}=-{\sum}_{m=1}^{M}\log P(\bm{x_m})
\end{aligned}
\end{equation}
where $M$ is the total number of masked tokens and $P(\bm{x_m})$ is the predicted probability of the token $x_m$ over the vocabulary size.

\textbf{Class Contrastive Discrimination.} To better discriminate different classes of entity representations in semantic space, we introduce class contrastive discrimination. Specifically, we construct positive (negative) samples as follows:

Given an input $x$ that contains $K$ entities, we employ the following procedure to generate positive and negative samples. For positive samples, we replace these $K$ entities with their corresponding label mentions to create $K$ positive samples for each input utterance. For negative samples, we select samples from other classes within the batch. Additionally, we replace all entities with irrelevant label mentions to construct a hard negative sample for each instance that is easily confused by the model. These hard negative samples are then included in the negative sample set. Figure \ref{fig:pretrain} illustrates the corresponding positive and negative samples as depicted in our experiment.

The representations of the original, positive, and negative samples are denoted by $h_{o}$, $h_{p}$, and $h_{n}$, respectively. To account for multiple positive samples, we adopt the supervised contrastive learning (SCL) objective \cite{khosla2020supervised}, which aims to minimize the distance between the original samples $h_{o}$ and their semantically similar positive samples $h_{p}$, while maximizing the distance between $h_{o}$  
and 2 samples: the negative samples $h_{n}$ and the hard negative samples $h_{hn}$. The formulation of $L_{SCL}$ is as follow:
\begin{equation}
\begin{split}
{\mathcal{L}_{SCL}} = &\frac{-1}{N}\sum_{i=1}^{N}\frac{1}{N_{y_i}} \sum_{j=1}^{N_{y_i}}\sum_{k=1}^{N_{y_i}} \mathbb I_{y_{ij} = y_{ik}}\\ &\log \frac{e^{sim(h_{oij}, h_{pik})/\tau}}{\sum_{l=1}^{N} (\mathbb I_{j \neq l})e^{sim(h_{oij}, h_{nl})/\tau}+ e^{sim(h_{oij}, h_{hnij})/\tau}} 
\end{split}
\end{equation}
where $N$ and $N_{y_i}$ denote the number of total examples in the batch and positive samples. $\tau$ is a temperature hyperparameter and $sim(h_1, h_2)$ is cosine similarity. \textbf{1} is an indicator function.

We sum the demonstration-base MLM task loss and the
class contrastive discrimination task loss, and finally obtain
the overall loss function L:
\begin{equation}
\begin{aligned}
L = \alpha L_{mlm}+ (1-\alpha) L_{SCL}
\end{aligned}
\end{equation}
where $L_{mlm}$ and $L_c$ denote the loss functions of the two
tasks. In our experiments, we set $\alpha = 0.6$.

\subsection{Downstream Few-shot NER}
After the pre-training stage, our model initially learns different aspects of information. In this section, We formally present the notations and the techniques of our proposed MSDP in the fine-tuning stage. Figure \ref{fig:main} illustrates the overall framework, which is composed of two steps: span extraction and entity classification.

\subsubsection{\textbf{Notations}}

We denote the train and test sets by $\mc D_{train}$ and $\mc D_{test}$, respectively. Both of them have the form of meta-learning datasets. The dataset consists of multiple episodes of data, and each episode of data $\mc E = (\mc S, \mc Q)$ consists of a support set $\mc S$ and a query set $\mc Q $. A sample $(X, \mc Y)$ in the support set or query set consists of the input sentence $X$ and the label set $\mc Y = \{(s_j, e_j, y_j)\}_{j=1}^N$ ($N$ denoting the number of spans,  $s_j$ and $e_j$ denoting the start and end positions of the $j$-th span, $y_j$ denoting the category of the $j$-th span).
$\bs h$ denotes the hidden representation obtained by encoding the input text.

\subsubsection{\textbf{Span Extractor}}
The span extractor aims to detect all entity spans. We initialize encoder with pre-trained parameters to encode the input sentence as a hidden representation $\bs h$, and calculate attention scores between each token representation to judge the start token and end token of the entity span. Following previous works \cite{ma2022decomposed,wang2022spanproto}, we use span-based cross-entropy as the loss function to optimise our encoder. We first design the weight matrixes $W_q/W_k/W_v$ of values $q/k/v$ and bias $b_q/b_k$ for the attention mechanism, and then compute the attention score of the $i$-th and $j$-th token, using formula: $f(i,j)=q_i^Tk_j+W_v(\bs h_i+\bs h_j)$. $\Omega_{i,j}$ indicates whether the span bounded by $i$, $j$ is an entity. Therefore, the span-based cross-entropy can be expressed as:
\begin{equation}
    \mathcal{L}_{span} = \log(1+  \hspace{-2mm} \sum_{1\leq i < j \leq L} 
  \hspace{-2mm} \exp((-1)^{\Omega_{i,j}}f(i,j))
\end{equation}

\subsubsection{\textbf{Entity Classification}}
In the second stage, we classify the entity spans extracted in the first stage. Different from the previous methods only computing the original prototype, we further decompose class-oriented prototypes and contextual fusion prototypes by two masking strategies, which introduce different information to assist in the classification task, thus alleviating the prototype classification disarray problem.

\textbf{1) Semantic masking strategies.} Firstly, we introduce two novel semantic masking strategies for the subsequent construction of semantic decomposing prototypes.

\begin{itemize}
\item {\textbf{Class-oriented Masking}}: Given an input sentence $X=\{x_1, x_2,\\ \dots, x_L\}$, we replace all the entity spans in $X$ whose labels that are not $y$ with [MASK] tokens to obtain class $y$ specific input $X_{\rm cs}^{y}$, thereby forcing the model to focus on the information of specific class by shielding the interference of other entities. For example, as shown in Figure \ref{fig:main}, we replace the “school” entity of the “LOC” class and the “Sunday” entity of the “DAY” class with [MASK] tokens to obtain “PER” class-oriented input\textit{ Mike wants to go to [MASK] on [MASK]. }

\item {\textbf{Contextual Fusion Masking}}: we replace all the entities in a sentence with [MASK] tokens, thus allowing the model to focus more on contextual fusion information. As the example sentence in Figure \ref{fig:main}, we mask all entities to obtain $X_{\rm ctx}=$ \textit{[MASK] wants to go to [MASK] on [MASK].}
\end{itemize}

\textbf{2) Prototype Constructing.} After decomposing two types of inputs with different information, we construct original prototype and two extra prototypes for each class in entity classification stage (The upper right corner of Figure \ref{fig:main})

For original prototype, we add up the representations of the start token and the end token of an entity span as the span boundary representation:
\begin{equation}
    \bs u_j = \bs h_{s_j}+\bs h_{e_j}
    \label{eq:uj}
\end{equation}
where $\bs u_j$ is the representation of the $j$-th span in the sentence, $\bs h_i$ denote the representation of the $i$-th token in the sentence. $s_j$ and $e_j$ are the start and end positions of the $j$-th span respectively.

For class-oriented prototype, we perform a class-oriented masking strategy for class $\bs t$ on $X$ to obtain $X^{\rm t}_{\rm cs}$, and compute a span representation $\bs u^{\rm cs}_j$ in $X^{\rm t}_{\rm cs}$ according to equation \ref{eq:uj}. 

For contextual fusion prototype, we perform all the entity-masking strategy on the original sentence $X$ to obtain $X_{\rm ctx}$ and then compute the span representation $\bs u^{\rm ctx}_j$ by averaging the representations of all tokens as follow:
\begin{equation}
    \bs u^{\rm ctx}_j = \frac{1}{L}\sum_{i=1}^L\bs h_i
\end{equation}
where $L$ denotes the number of tokens in $X_{\rm ctx}$.

Afterwards, we construct three different prototypes vectors by averaging the representations of all entities of the same class in the support set:
\begin{equation}
    \bs c_t = \frac{\sum_{(X,\mc Y)\in \mathcal S}\sum_{(s_j,e_j,y_j)\in \mc Y}\mathbb I(y_j=t)\bs u}{\sum_{(X,\mc Y)\in \mathcal S}\sum_{(s_j,e_j,y_j)\in \mc Y}\mathbb I(y_j=t)}
\end{equation}
where $\mathbb I(\cdot) $ is the indicator function; $\bs u$ can be replaced with $\bs u_j$, $\bs u^{\rm cs}_j$and $ \bs u^{\rm ctx}_j$to calculate three semantic prototypes separately.

After constructing three different semantic prototypes, we use a metric-based approach for classification and optimize the parameters of the model based on the basis of distance between entity representations and class prototypes:
\begin{equation}
    \mathcal L_{cls} = \sum_{(X,\mc Y)\in\mc S}\sum_{(s_j,e_j,y_j)\in \mc Y}
    -\log p(y_j|s_j, e_j)
\end{equation}
where 
\begin{equation}
    p(y_j|s_j, e_j) = {\rm softmax}(-d(\bs u_j, \bs c_{y_j}))
\end{equation}
is the probability distribution. We use cosine similarity as the distance function $d(\cdot, \cdot)$.

\subsection{Training and Inference of MSDP}
We first perform two task-specific pre-trainings to learn reliable entity boundary information and entity representations of different classes. For fine-tuning, we initialize the BERT encoder with pre-trained parameters for the few-shot NER task.

\begin{table}[htbp]
  \renewcommand\tabcolsep{3pt}
  \centering
    \begin{tabular}{cccc}
    \hline
    \textbf{Dataset} & \textbf{Domain} & \textbf{\# Sentences} & \textbf{\# Classes} \bigstrut\\
    \hline
    Few-NERD & Wikipedia & 188.2k & 66 \bigstrut[t]\\
    CoNLL03 & News  & 20.7k & 4 \\
    GUM   & Wiki  & 3.5k  & 11 \\
    WNUT  & Social & 5.6k  & 6 \\
    OntoNotes & Mixed & 159.6k & 18 \bigstrut[b]\\
    \hline
    \end{tabular}%
  \caption{Evaluation dataset statistics.}
  \label{tab:datasets}%
  \vspace{-0.2cm}
\end{table}%
In the downstream training phase, given the training set $\mc D_{train} = { (\mc S, \mc Q) }$, we compute $\mc L_{span}$ and three types of prototypes on the support set $\mc S$, $\mc L_{cls}$ on the query set $\mc Q$, and train the two optimization objectives jointly. Following SpanProto \cite{wang2022spanproto}, we only optimize the $\mc L_{span}$ objective in the first $T$ steps, and jointly optimize both $\mc L_{span}$ and $\mc L_{cls}$ after $T$ steps. 

In the testing phase, given an episode $\mc E=(\mc S, \mc Q) \in \mc D_{test}$, we construct prototypes on the support set $\mc S$ and perform span detection on the query set $\mc Q$. Then we calculate the distance between the extracted spans and each class prototype for classification. Note that we only utilize original prototypes during inference.

\section{Experiment}
\label{sec:page}

\subsection{Datasets}
Table \ref{tab:datasets} shows the dataset statistics of original data for constructing few-shot episodes.
We evaluate our method on two widely used few-shot benchmarks Few-NERD \cite{ding2021few} and CrossNER \cite{liu2021crossner}. 

\textbf{Few-NERD}: Few-NERD is annotated with 8 coarse-grained and 66 fine-grained entity types, which consists of two few-shot settings (Intra, and Inter). In the Intra setting, all entities in the training set, development set, and testing set belong to different coarse-grained types. In contrast, in the Inter setting, only the fine-grained entity types are mutually disjoint in different datasets. we use episodes released by \citeauthor{ding2021few} which contains 20,000 episodes for training, 1,000 episodes for validation, and 5,000 episodes for testing. Each episode is an $N$-way $K\sim 2K$-shot few-shot task.

\textbf{CrossNER}: CrossNER contains four domains from CoNLL-2003 \cite{sang2003introduction}(News), GUM \cite{zeldes2017gum} (Wiki), WNUT-2017 \cite{derczynski2017results} (Social), and Onto-\\notes \cite{pradhan2013towards}(Mixed). We randomly select two domains for training, one for validation, and the remaining for testing. We use public episodes constructed by \citeauthor{hou2020few} .


\begin{table*}[htbp]
  \centering
  \resizebox{1\linewidth}{!}{
    \begin{tabular}{cl|ccccc|ccccc}
    \toprule
    \multicolumn{2}{l|}{\multirow{2}[4]{*}{\textbf{Paradigms Models}}} & \multicolumn{5}{c|}{\textbf{Intra}}   & \multicolumn{5}{c}{\textbf{Inter}} \\
\cmidrule{3-12}    \multicolumn{2}{c|}{} & \multicolumn{2}{c}{\textbf{1$\sim$2-shot}} & \multicolumn{2}{c}{\textbf{5$\sim$10-shot}} & \multirow{2}[2]{*}{Avg.} & \multicolumn{2}{c}{\textbf{1$\sim$2-shot}} & \multicolumn{2}{c}{\textbf{5$\sim$10-shot}} & \multirow{2}[2]{*}{Avg.} \\
    \multicolumn{2}{c|}{} & 5 way & 10 way & 5 way & 10 way &       & 5 way & 10 way & 5 way & 10 way &  \\
    \midrule
    \multirow{4}[2]{*}{\textit{One-stage}} & ProtoBERT & $23.45_{ \pm{0.92}}$ & $19.76_{ \pm{0.59}}$ & $41.93_{ \pm{0.55}}$ & $34.61_{ \pm{0.59}}$ & 29.94 & $44.44_{ \pm{0.11}}$ & $39.09_{ \pm{0.87}}$ & $58.80_{ \pm{1.42}}$ & $53.97_{ \pm{0.38}}$ & 49.08 \\
          & NNShot & $31.01_{ \pm{1.21}}$ & $21.88_{ \pm{0.23}}$ & $35.74_{ \pm{2.36}}$ & $27.67_{ \pm{1.06}}$ & 29.08 & $54.29_{ \pm{0.40}}$ & $46.98_{ \pm{1.96}}$ & $50.56_{ \pm{3.33}}$ & $50.00_{ \pm{0.36}}$ & 50.46 \\
          & StructShot & $35.92_{ \pm{0.69}}$ & $25.38_{ \pm{0.84}}$ & $38.83_{ \pm{1.72}}$ & $26.39_{ \pm{2.59}}$ & 31.63 & $57.33_{ \pm{0.53}}$ & $49.46_{ \pm{0.53}}$ & $57.16_{ \pm{2.09}}$ & $49.39_{ \pm{1.77}}$ & 53.34 \\
          & CONTaiNER & 40.43 & 33.84 & 53.70  & 47.49 & 43.87 & 55.95 & 48.35 & 61.83 & 57.12 & 55.81 \\
    \midrule
    \multirow{4}[2]{*}{\textit{Two-stage}} & ESD   & $41.44_{ \pm{1.16}}$ & $32.29_{ \pm{1.10}}$ & $50.68_{ \pm{0.94}}$ & $42.92_{ \pm{0.75}}$ & 41.83 & $66.46_{ \pm{0.49}}$ & $59.95_{ \pm{0.69}}$ & $74.14_{ \pm{0.80}}$ & $67.91_{ \pm{1.41}}$ & 67.12 \\
          & DecomMeta & $52.04_{ \pm{0.44}}$ & $43.50_{ \pm{0.59}}$ & $63.23_{ \pm{0.45}}$ & $56.84_{ \pm{0.14}}$ & 53.9  & $68.77_{ \pm{0.24}}$ & $63.26_{ \pm{0.40}}$ & $71.62_{ \pm{0.16}}$ & $68.32_{ \pm{0.10}}$ & 67.99 \\
          & SpanProto & $54.49_{ \pm{0.39}}$ & $45.39_{ \pm{0.72}}$ & $65.89_{ \pm{0.82}}$ & $59.37_{ \pm{0.47}}$ & 56.29 & $73.36_{ \pm{0.18}}$ & $66.26_{ \pm{0.33}}$ & $75.19_{ \pm{0.77}}$ & $70.39_{ \pm{0.63}}$ & 71.3 \\
          & \textbf{MSDP} & \bm{$56.35_{ \pm{0.28}}$} & \bm{$47.13_{ \pm{0.69}}$} & \bm{$66.80_{ \pm{0.78}}$} & \bm{$64.69_{ \pm{0.51}}$} & \textbf{58.74} & \bm{$76.86_{ \pm{0.22}}$} & \bm{$69.78_{ \pm{0.31}}$} & \bm{$84.78_{ \pm{0.69}}$} & \bm{$81.50_{ \pm{0.71}}$} & \textbf{78.23} \\
    \bottomrule
    \end{tabular}}
    \caption{F1 scores with standard deviations on Few-NERD for both inter and intra settings.}
    \label{tab:table1}%
\end{table*}%

\begin{table*}[htbp]
  \centering
   \resizebox{1\linewidth}{!}{
    \begin{tabular}{cl|llllr|llllr}
     \toprule
    \multicolumn{2}{l|}{\multirow{1}[3]{*}{\textbf{Paradigms Models}}} & \multicolumn{5}{c|}{\textbf{1-shot}}  & \multicolumn{5}{c}{\textbf{5-shot}} \\
\cmidrule{3-12}   \multicolumn{2}{c|}{} & \multicolumn{1}{c}{CONLL-03} & \multicolumn{1}{c}{GUM} & \multicolumn{1}{c}{WNUT-17} & \multicolumn{1}{c}{OntoNotes} & \multicolumn{1}{c|}{Avg.} & \multicolumn{1}{c}{CONLL-03} & \multicolumn{1}{c}{GUM} & \multicolumn{1}{c}{WNUT-17} & \multicolumn{1}{c}{OntoNotes} & \multicolumn{1}{c}{Avg.} \\
    \midrule
    \multirow{3}[2]{*}{\textit{One-stage}} & Matching Network & $19.50_{ \pm{0.35}}$ & $4.73_{ \pm{0.16}}$ & $17.23_{ \pm{2.75}}$ & $15.06_{ \pm{1.61}}$ & 14.13 & $19.85_{ \pm{0.74}}$ & $5.58_{ \pm{0.23}}$ & $6.61_{ \pm{1.75}}$ & $8.08_{ \pm{0.47}}$ & 10.03 \\
          & ProtoBERT & $32.49_{ \pm{2.01}}$ & $3.89_{ \pm{0.24}}$ & $10.68_{ \pm{1.40}}$ & $6.67_{ \pm{0.46}}$ & 13.43 & $50.06_{ \pm{1.57}}$ & $9.54_{ \pm{0.44}}$ & $17.26_{ \pm{2.65}}$ & $13.59_{ \pm{1.61}}$ & 22.61 \\
          & L-TapNet+CDT & $44.30_{ \pm{3.15}}$ & $12.04_{ \pm{0.65}}$ & $20.80_{ \pm{1.06}}$ & $15.17_{ \pm{1.25}}$ & 23.08 & $45.35_{ \pm{2.67}}$ & $11.65_{ \pm{2.34}}$ & $23.30_{ \pm{2.80}}$ & $20.95_{ \pm{2.81}}$ & 25.31 \\
    \midrule
    \multirow{3}[2]{*}{\textit{Two-stage}} & DecomMeta & $46.09_{ \pm{0.44}}$ & $17.54_{ \pm{0.98}}$ & $25.14_{ \pm{0.24}}$ & $34.13_{ \pm{0.92}}$ & 30.73 & $58.18_{ \pm{0.87}}$ & $31.36_{ \pm{0.91}}$ & $31.02_{ \pm{1.28}}$ & $45.55_{ \pm{0.90}}$ & 41.53 \\
          & SpanProto & $47.70_{ \pm{0.49}}$ & $19.92_{ \pm{0.53}}$ & $28.31_{ \pm{0.61}}$ & $36.41_{ \pm{0.73}}$ & 33.09 & $61.88_{ \pm{0.83}}$ & $35.12_{ \pm{0.88}}$ & $33.94_{ \pm{0.50}}$ & $48.21_{ \pm{0.89}}$ & 44.79 \\
          & \textbf{MSDP} & \bm{$49.14_{ \pm{0.52}}$} & \bm{$21.88_{ \pm{0.29}}$} & \bm{$30.10_{ \pm{0.56}}$} &\bm{$38.05_{ \pm{0.88}}$} & \textbf{34.79} & \bm{$63.98_{ \pm{0.80}}$} & \bm{$36.53_{ \pm{0.81}}$} & \bm{$35.61_{ \pm{0.72}}$} & \bm{$49.99_{ \pm{0.95}}$} & \textbf{46.53} \\
    \bottomrule
    \end{tabular}}
  \caption{F1 scores with standard deviations under 1 shot and 5 shot setting on CrossNER.}
    \label{tab:table2}%
\end{table*}%
\subsection{Baselines}
For the baselines, we choose multiple strong approaches from the paradigms of one-stage and two-stage. 1) One-stage NER paradigms: ProtoBERT \cite{snell2017prototypical}, StructShot \cite{yang2020simple}, NNShot \cite{yang2020simple}, CONTAINER \cite{das2021container} and LTapNet+CDT \cite{hou2020few}. 2) Two-stage paradigm: ESD \cite{wang2021enhanced}, MAML-ProtoNet \cite{ma2022decomposed} and SpanProto \cite{wang2022spanproto}. Due to the space limitation, More details of these baselines and implementations are illustrated as follow:
\begin{itemize}
\item {\textbf{SimBERT}} \cite{hou2020few} applies BERT
without any finetuning as the embedding function, then assigns each token’s label by retrieving the most similar token in the support set.

\item {\textbf{ProtoBERT}} \cite{fritzler2019few} uses a token-
level prototypical network \cite{snell2017prototypical} which represents each class by averaging token representations with the same label, then the label of each token in the query set is decided by its nearest class prototype.

\item {\textbf{MatchingBERT}} \cite{NIPS2016_90e13578} is similar to ProtoBERT except that it calculates the similarity between query instances and support instances instead of class prototypes.

\item {\textbf{L-TapNet+CDT}} \cite{hou2020few} enhances
TapNet with pair-wise embedding, label semantic, and CDT transition mechanism.

\item {\textbf{NNShot}} \cite{yang2020simple} pretrains
BERT for token embedding by conventional classification for training, and a token-level nearest neighbor method is used at testing. 

\item {\textbf{StructShot}} \cite{yang2020simple} improves
NNshot by using an abstract transition probability for Viterbi decoding at testing.

\item {\textbf{ESD}} \cite{wang2021enhanced} is a span-level metric learning-based method. It enhances the prototypical network by using inter- and cross-span attention for better span representation and designs multiple prototypes for O label. 

\item {\textbf{TransferBERT}} \cite{hou2020few} trains a token-
level BERT classifier, then finetunes task-specific linear classifier on the support set at test time.

\item {\textbf{CONTAINER}} \cite{das2021container} uses token-
level contrastive learning for training BERT as a token embedding function, then finetunes the BERT on the support set and applys the nearest neighbor method at inference time.

\item {\textbf{DecomMeta}} \cite{ma2022decomposed} trains the span detector by introducing the model-agnostic meta-learning (MAML) algorithm and uses MAML-enhanced prototypical networks to find a good embedding space.

\item {\textbf{SpanProto}} \cite{wang2022spanproto} transforms the sequential tags into a global boundary matrix in the span extraction stage and performs prototypical learning with a margin-based loss in the mentioned classification stage.

\end{itemize}

\subsection{Implementation Detail}
\label{sec:implement}
For the upstream work, we use BERT-base-uncased \cite{devlin2018bert} from HuggingFace as the backbone. In two pre-training settings, we set the batch size of BERT to 8 and the pre-training takes an average of 12 hours for 5 epochs. The corresponding learning rates are set to 1e-5. We set the number (K) of the retrieved demonstrations to 5 and the negative demonstration to 3. We set temperature hyperparameter$\tau_1$ to 0.5. In addition, our upstream pre-training corpus is aligned with the downstream task, which means no additional data will be introduced in the pre-training stage. For instance, if the downstream few-shot NER experiment is conducted on the Inter 5 way 1-2 shot of Few-NERD, the pre-training data is the training set of Inter 5 way 1-2 shot.

For the downstream work, we adopt the standard N-way K-shot setting \cite{ding2021few} and align the task definition with previous work \cite{ma2022decomposed}. We choose Adam \cite{kingma2014adam} as the optimizer with a learning rate of 3e-5. The warm-up rate is set to 0.1. The max sequence length we set is 64 and the batch size is set to 4. The training steps T and T' are set as 2000 and 200, respectively. For all the experiments, we train and test our model on the 3090Ti GPU. It takes an average of 5 hours to run with 3 epochs on the training dataset.

All experiments are repeated three times with different random seeds under the same settings. All the models are implemented with PyTorch. We will release our code after blind review.

\subsection{Main Results}

Table \ref{tab:table1} and Table \ref{tab:table2} report the main results compared with other baselines. We conduct the following comparison and analysis:
1) Our proposed method significantly outperforms all the previous methods in different settings. Specifically, compared with SpanProto, MSDP achieves a performance improvement on the overall averaged results over Few-NERD Intra by 4.3\% and Inter by 9.7\%. Meanwhile, MSDP shows a 3.9\% increase on CrossNER. Both results demonstrate the effectiveness of MSDP.
2) All methods in the two-stage paradigm perform better than those one-stage methods, which demonstrates the framework advantages of the span-based approach.
3) The overall performance of the Inter scenario is higher than Intra, since all entities in the training set/development set/test set belong to different coarse-grained types in the Intra setting. We still obtain extraordinary improvement in this challenging situation. 
All the results show that MSDP can adapt to a new domain in which the coarse-grained and fine-grained entity types are both unseen, which highlights the strong transferring ability of our approach.

\subsection{Ablation Studies. }We conduct an ablation study to investigate the characteristics of the main components in MSDP. Table \ref{tab:table3} shows the ablation results, and “w/o" denotes the model performance without a specific module. We have following observations: 1) The performance of MSDP drops when removing any one component, which suggests that every part of the design is necessary 2) Removing any one semantic prototype results in great performance degradation. This is consistent with our conjecture since class-oriented prototypes and contextual fusion prototypes provide relatively orthogonal semantic information from two perspectives. Missing each part will make the semantic space more chaotic and make the classification effect worse. 3) Removing joint pre-training tasks causes obvious performance degradation compared with removing one of them, which indicates that jointly pre-training objectives have a mutually reinforcing effect.


\begin{table}[!tbp]
\centering
  \resizebox{1\linewidth}{!}{
  \begin{tabular}{l|cccc}
    \toprule
    \multirow{2}[2]{*}{\textbf{Methods}} & \multicolumn{2}{c}{\textbf{Few-NERD}} & \multicolumn{2}{c}{\textbf{CrossNER}} \\
          & Intra & Inter & 1-shot & 5-shot \\
    \midrule
    \textbf{MSDP} & \textbf{58.49} & \textbf{78.23} & \textbf{34.79} & \textbf{46.53} \\
    \midrule
    w/o Demonstration-based MLM & 54.84 & 75.25 & 32.57 & 44.77 \\
    w/o Class Contrastive Discrimination & 56.97 & 74.87 & 31.66 & 43.54 \\
    w/o class-oriented Prototype & 55.04 & 74.58 & 33.14 & 44.48 \\
    w/o contextual fusion Prototype & 56.53 & 76.15 & 32.98 & 45.08 \\
    \midrule
    w/o Joint pre-training tasks & 53.58 &73.03&  30.30 & 41.51\\
    w/o Both two semantic prototypes & 54.32 &73.32 &31.74 & 43.22\\

    \bottomrule
    \end{tabular}%
    }
    \caption{The ablation study results (average F1 score
\%) for Few-NERD and CrossNER. }
  \label{tab:table3}%
\end{table}%


\begin{table}[t]
  \centering
   \resizebox{1\linewidth}{!}{
      \begin{tabular}{l|cccc|cccc}
    \toprule
    \multirow{3}[4]{*}{\textbf{Method}} & \multicolumn{4}{c|}{\textbf{Inter(1-2shot)}} & \multicolumn{4}{c}{\textbf{Intra(1-2shot)}} \\
\cmidrule{2-9}          & \multicolumn{2}{c}{5 way} & \multicolumn{2}{c|}{10 way} & \multicolumn{2}{c}{5 way} & \multicolumn{2}{c}{10 way} \\
          & Pre. & Rec. & Pre & Rec. & Pre. & Rec.& Pre. & Rec. \\
    \midrule
    MSDP (Base)& 71.6  & 100   & 75.5 & 100   & 72.8 & 100   & 73.7 & 100 \\
    + Pre-training & 74.3 & 100   & 77.5 & 100   & 73.9 & 100   & 74.8 & 100 \\
    \midrule
    MSDP (Full) & \textbf{75.2}  & 100   & \textbf{78.1 }& 100   & \textbf{74.3 } & 100   & \textbf{75.3}  & 100 \\
    \bottomrule
    \end{tabular}%
    }
    \caption{The span extractor performance (average Precision and Recall) on Few-NERD 1-2 shot. MSDP(Base) denotes MSDP without two pre-training and prototypes.}
  \label{tab:table4}%
\end{table}%

\begin{figure}[t]
\centering
\resizebox{.48\textwidth}{!}{\includegraphics{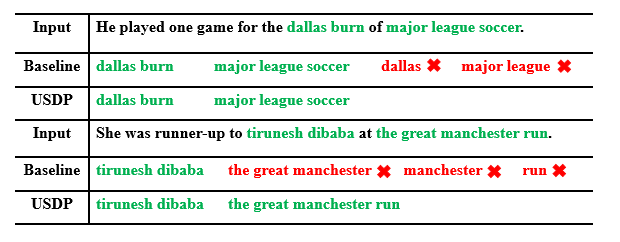}}
\vspace{-0.2cm} 
\caption{The cases of Few-NERD. Both wrong and correct labels are marked in red and green, respectively. }

\label{fig:case}
\vspace{-0cm} 
\end{figure}

\subsection{Effectiveness on Span Over-prediction}

\textbf{Qualitative analysis.} Span over-prediction causes the model to extract redundant candidate spans in addition to predicting the correct spans.  This phenomenon can be reflected in high recall rate and low precision rate of the span extractor. As shown in Table \ref{tab:table4}, compared with the MSDP(Base), joint pre-training tasks improve the prediction accuracy(2.7\% for 5 way and 2.0\% for 10 way) while maintaining a high recall rate, which proves that joint pre-training tasks can bring entity boundary information and better representation into PLMs. For MSDP(Full), we unexpectedly find that the semantic decomposing method also improves the precision rate slightly. Since the joint training of span extractor and entity classification, both contextual fusion and class-oriented information also have a positive effect on distinguishing entity boundaries.

\textbf{Case Study for Span Extractor} To further verify the effect of our MSDP on Span Over-prediction, we randomly sample 100 instances from outputs and select two representative cases in figure \ref{fig:case}. The baseline model even generates some wrong spans in order to predict all spans while our method does not require such a cost. These cases suggest that MSDP captures more reliable entity-boundary information. In summary, we demonstrate that MSDP can better solve the over-prediction problem from both statistical and sample aspects.

\nocite{Ando2005,augenstein-etal-2016-stance,andrew2007scalable,rasooli-tetrault-2015,goodman-etal-2016-noise,harper-2014-learning}
\subsection{Performance on  Classification Disarray}
\textbf{Error Analysis} We follow \cite{wang2022spanproto} to conduct error analysis in Table \ref{tab:table5}. Results show that MSDP outperforms other strong baselines with fewer false positive prediction errors. Specifically, we achieve 9.22\% of “FP-Type” when getting 76.86 F1 scores. Meanwhile, this suggests our MSDP obtains the lowest error rate and effectively solves the problem of prototype classification disarray.

\begin{figure}[t]
    \centering
    \subfigure[MSDP(Base)]{
        \includegraphics[width=.22\textwidth]{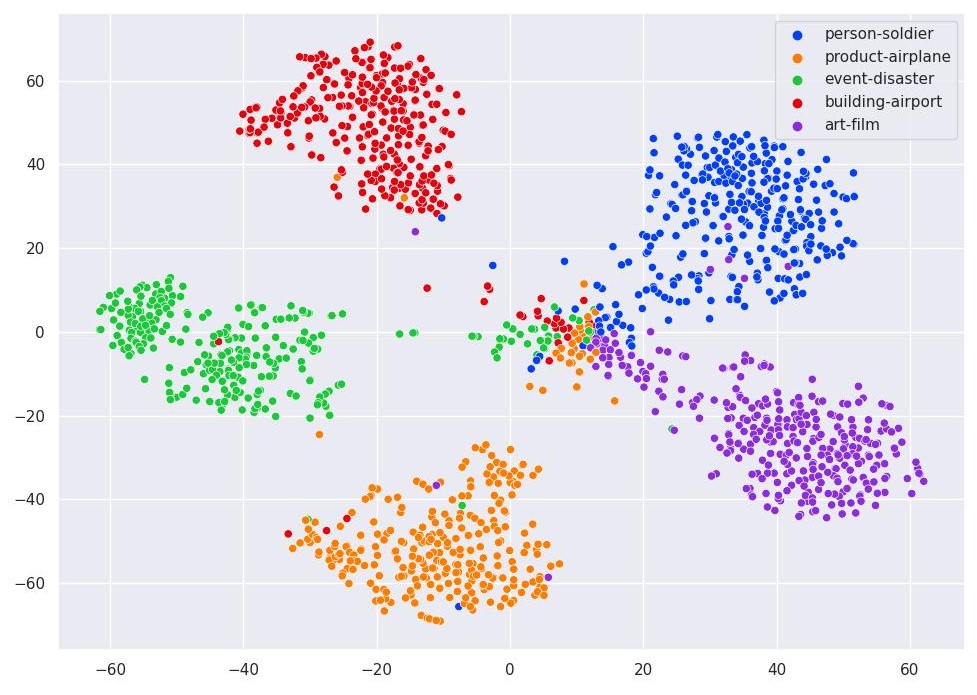}
        \label{label_for_cross_ref_1}
    }
    \subfigure[MDSP(w. Pre-training)]{
	\includegraphics[width=.22\textwidth]{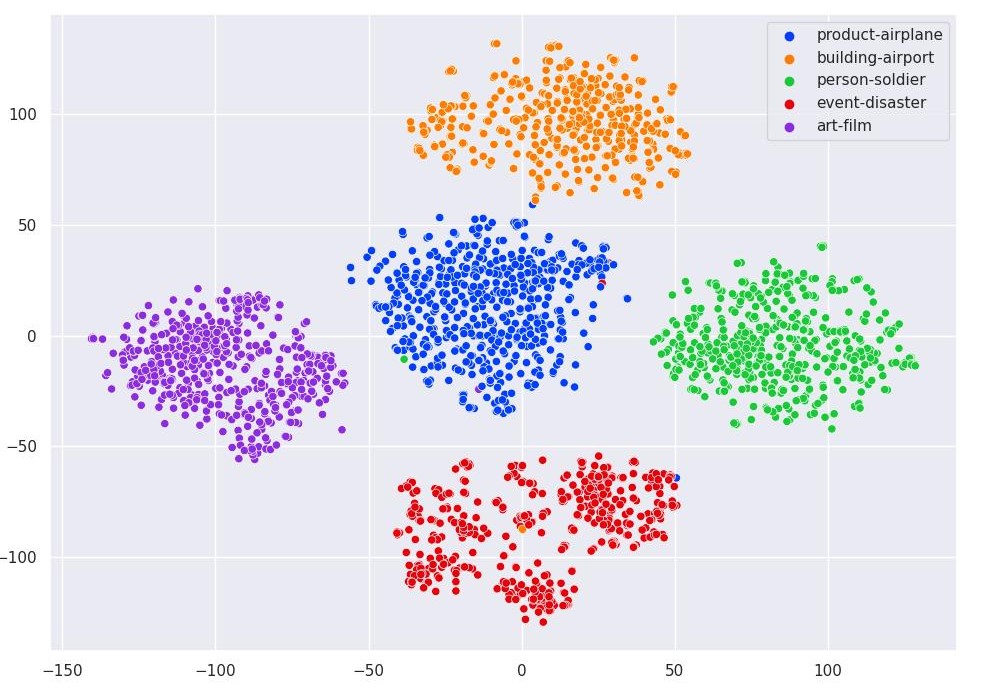}
        \label{label_for_cross_ref_2}
    }
    \quad    
    \subfigure[MSDP (Full)]{
        \includegraphics[width=.22\textwidth]{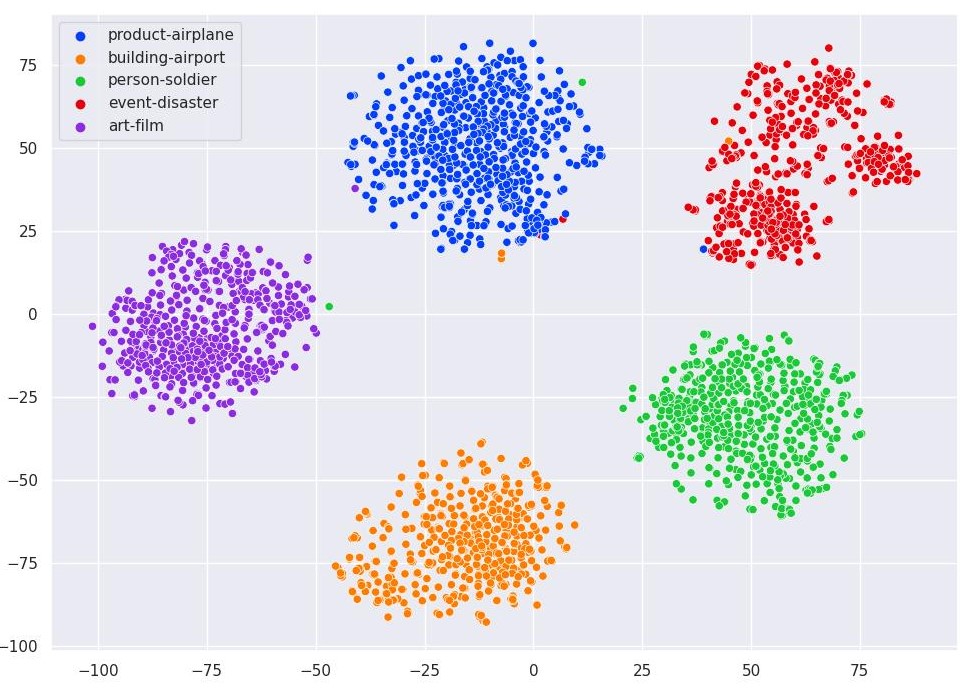}
        \label{label_for_cross_ref_3}
    }
    \caption{The t-SNE visualization of the span representations with 500 5-way 1-shot data from Few-NERD Inter for both SpanProto and MSDP. The points with different colors denote the entity span with different types. }
    \label{fig:visual}
    \vspace{-0.2cm}
\end{figure}

\textbf{Visualization} To further explore the effectiveness of MSDP on prototype classification disarray problems, we investigate how our MSDP adjusts the representations in the semantic space. We use 500 5-way 1-shot episodes data from Few-NERD Inter for training, and visualize the span representations of 4 types of entity by t-SNE toolkit \cite{van2008visualizing} in three different settings: MSDP(Base), MSDP(with Pre-training) and MSDP(Full). As shown in Figure \ref{fig:visual}, the span representations of each class are gathered around the corresponding type prototype region. Compared with MSDP(Base), joint pre-training tasks help the model increases the distance between the representations of different classes. For the MSDP(full), the intra-class distance is further compressed due to the optimization of the semantic decomposing method. In this way, both pre-training and decomposing methods improve the quality of entity representations from different aspects. Thus, MSDP effectively alleviates the prototype classification disarray problem in the entity classification stage.

\subsection{Influence of Data Size}

\begin{figure}[t]
\centering
\resizebox{.47\textwidth}{!}{\includegraphics{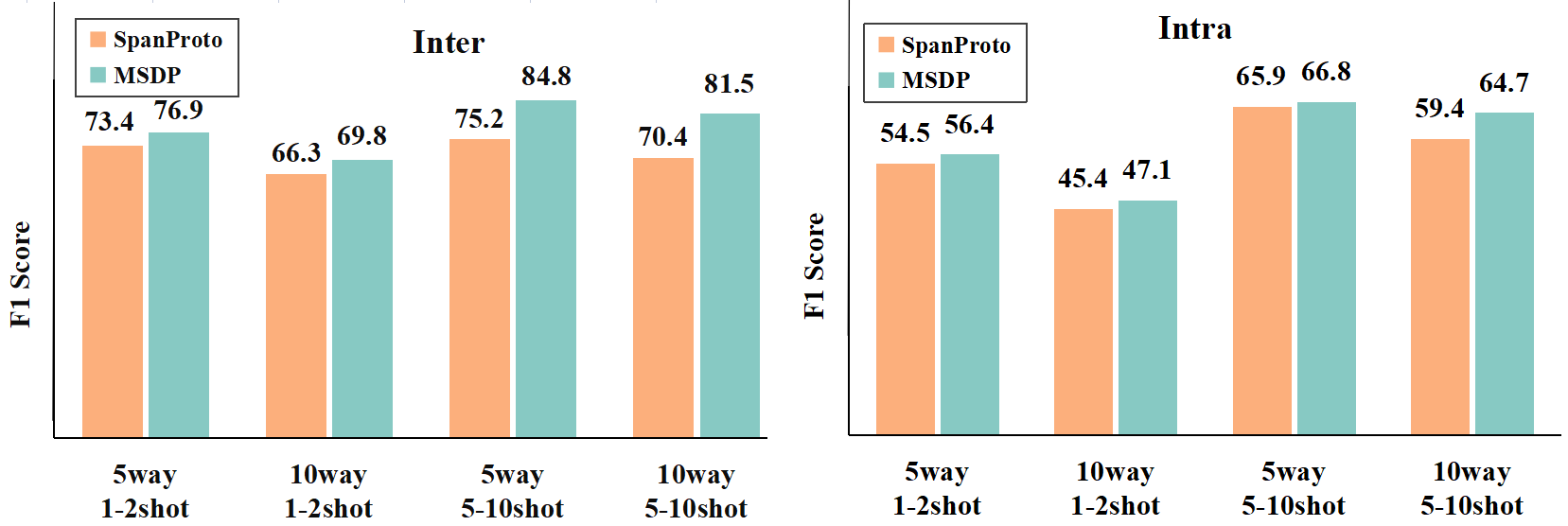}}
\caption{The influence of data size under Inter(left) and Intra(right) setting of Few-NERD}
\label{fig:6}
\end{figure}

To find out the influence of data size, we conduct a comparison experiment between SpanProto and MSDP under different few-shot settings of Few-NERD. As shown in figure \ref{fig:6}, the performance of MSDP still has a steady improvement compared with SpanProto with the increase of data in inter setting while the improvement is not obvious in intra setting. We think the possible reasons are as follows: since fine-grained entity types are separated in inter setting, compared with baseline, MSDP can better assist the model to capture the fine-grained entity type information, and the effect is more obvious with the increase of data size. For intra settings which are separated fine-grained entity types, the MSDP has a slight increase, but there is still much room for improvement. So improving the ability of model to capture coarse-grained entity types is a great challenge for future research.

\begin{table}[!tbp]
\centering
  \resizebox{0.5\linewidth}{!}{
  \begin{tabular}{l|cc}
    \toprule
    \multirow{1}[2]{*}{\textbf{Methods}} & \multicolumn{1}{c}{\textbf{F1}} & \multicolumn{1}{c}{\textbf{FP-Type}}  \\
    \midrule
    ProtoBERT & 44.44 & 13.30  \\
    NNShot & 54.29 & 15.30  \\
    StructShot & 57.33 & 20.00 \\
    ESD & 66.46 & 27.20  \\
    DecomMeta &  76.11 & 46.53 \\
    SpanProto &  73.36 & 10.90 \\
    \textbf{MSDP} & \textbf{76.86} & \textbf{9.22} \\

    \bottomrule
    \end{tabular}%
    }
    \caption{Error analysis (\%) of 5-way 1-shot on Few-NERD Inter.  “FP-Type” represents extracted entities with the right span boundary but the wrong entity type. }
  \label{tab:table5}%
\end{table}%


\section{Hyper-parameter Analysis}
\subsection{The Effect of Temperature Parameter.} Table \ref{tab:temperature} shows the effect of different $\tau$ values of SCL in class contrastive discrimination task. We find that when the temperature $\tau = 0.5$, MSDP achieves the best performance in the Inter and Intra setting of Few-NERD. Our method within $\tau$ [0.1, 0.6] outperforms sota baselines (SpanProto), and $\tau$ in [0.4, 0.6] brings larger improvements(above 2\% in Inter and 6\% in Intra).This experiment demonstrates the robustness of MSDP, as changes in temperature $\tau$ do not affect its performance.

\subsection{The Effect of Number of Demonstrations.} 
\begin{figure}[t]
\centering
\resizebox{.4\textwidth}{!}{\includegraphics{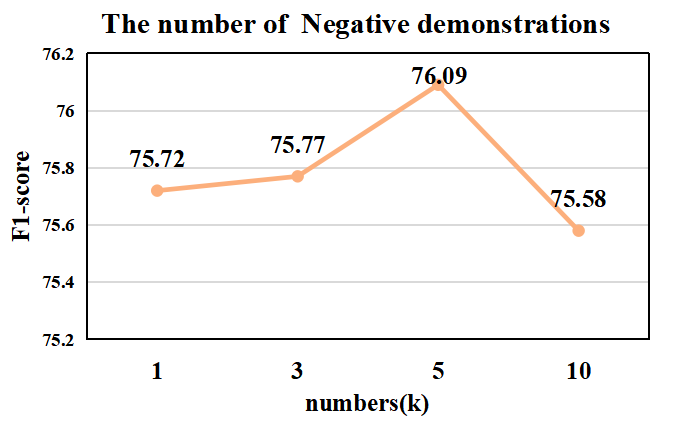}}
\resizebox{.4\textwidth}{!}{\includegraphics{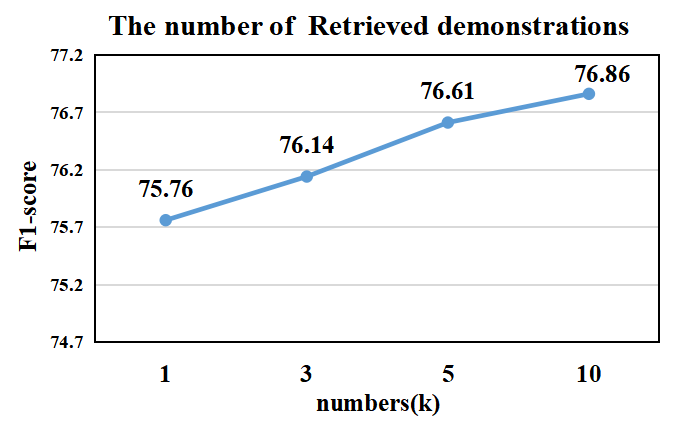}}
\caption{The performance of MSDP changes over the number(K) of retrieved demonstration and negative demonstration }
\label{fig:7}
\end{figure}

We further examine whether the performance of MSDP changes over the number(K) of retrieved demonstration and negative demonstration in the pre-training stage. As shown in Figure \ref{fig:7}, the performance of MSDP improves from 75.76 to 76.86 on Few-NERD Inter5-1 with the number of retrieved demonstrations from 1 to 10. However, the performance of negative demonstrations increases first(75.72 to 76.09) and then decreases(76.09 to 75.58) due to the increase in the number of demonstrations. The possible reason is that the introduction of retrieved demonstrations can provide rich entity-label pairs and factual information, which can assist the model to learn good representations. A small amount of negative demonstrations can help the model distinguish the boundary between entities and non-entities, but too many negative demonstrations will introduce a large number of non-entities which brings the noise.

\begin{table}[!tbp]
\small
\centering
  \resizebox{0.5\linewidth}{!}{
    \begin{tabular}{l|cc}
    \toprule
    \multicolumn{1}{c|}{\multirow{2}[2]{*}{\textbf{Temperature} $\tau$}} & \multicolumn{2}{c}{Few-NERD}  \\
     & \multicolumn{1}{c}{Intra} & \multicolumn{1}{c}{Inter} \\
    \midrule
    $\tau=0.1$   & 56.84     &  76.33     \\
    $\tau=0.2$  &  58.03    & 77.25   \\
    $\tau=0.3$  & 57.44     & 76.62    \\
    $\tau=0.4$   & 58.53     & 77.89   \\
    $\tau=0.5$  & \textbf{58.74}     & \textbf{78.23}    \\
    $\tau=0.6$ & 58.60     & 78.03    \\
    \bottomrule
    \end{tabular}%
    }
   \caption{The parameter analysis of the temperature hyperparameter $\tau$. }
  \label{tab:temperature}%
  \vspace{-0.2cm}
\end{table}%

\section{Conclusion}

In this paper, we propose a \textbf{M}ulti-Task \textbf{S}emantic \textbf{D}ecomposition Framework via Joint Task-specific \textbf{P}re-training (MSDP) for few-shot NER. Specifically, We introduce two novel pre-training tasks, Demonstration-based MLM and Class Contrastive Discrimination, to solve the span over-prediction and prototype classification disarray problem. Further, We design a multi-task joint optimization framework, and decompose class-oriented prototypes and contextual fusion prototypes to integrate two different semantic information for entity classification. Experimental results demonstrate that MSDP outperforms the previous SOTA methods in terms of overall performance. Extensive analysis further validates the effectiveness and generalization of our approach.

\bibliographystyle{ACM-Reference-Format}
\balance
\bibliography{acmart,anthology}










\end{document}